\documentclass{article}

\usepackage[utf8]{inputenc} 
\usepackage[T1]{fontenc}    
\usepackage{hyperref}       
\usepackage{url}            
\usepackage{booktabs}       
\usepackage{amsfonts}       
\usepackage{nicefrac}       
\usepackage{microtype}      
\usepackage{xcolor}         
\usepackage{amsmath}
\usepackage{bbm}
\usepackage{amssymb,mathtools}
\usepackage{multicol}
\usepackage{natbib}
\title{Learning representations by forward-propagating errors}

\author{%
Ryoungwoo Jang\\
Coreline Soft\\
Seoul, South Korea\\
  \texttt{jryoungw2035@gmail.com} \\
}

\begin{document}

\maketitle

\begin{abstract}
Back-propagation (BP) is widely used learning algorithm for neural network optimization. However, BP requires enormous computation cost and is too slow to train in central processing unit (CPU). Therefore current neural network optimizaiton is performed in graphical processing unit (GPU) with compute unified device architecture (CUDA) programming. In this paper, we propose a light, fast learning algorithm on CPU that is fast as CUDA acceleration on GPU. This algorithm is based on forward-propagating method, using concept of dual number in algebraic geometry. 
\end{abstract}
\section{Introduction}

In 1986, Rumelhart, Hinton, Williams suggested learning algorithm of the neural network, which is now usually called as back-propagation (BP) \cite{rumelhart1986learning}. Since then, deep neural networks became trainable algorithm and is prospered by AlexNet \cite{krizhevsky2017imagenet}. Uncountable researches have been proposed to train more accurate models, analyze model behavior, and enumerous fields. However, there is one profound question: Is the learning rule for neural network unique? 

It seems that Geoffrey Hinton have contemplated this problem for a long time. In a paper \cite{lillicrap2020backpropagation}, Hinton and his colleagues approached issues of backpropagation in various perspectives. In 2022, Hinton have suggested a new learning rule named forward-forward algorithm \cite{hinton2022forward}.

This paper approaches learning rule for neural network in a different perspective. In this paper, a mathematically sound concept, named as dual number is introduced, and develop a learning rule for dual number system.

\section{Mathematical Preliminaries}
Let $G$ be a set with binary operation, denoted by $\ast$. We abbreviate this set as $(G,\ast)$. Now if $(G,\ast)$ satisfies following conditions, we call $(G,\ast)$ as group.
\begin{enumerate}
    \item For all $g,h,k\in G$, $(g\ast h)\ast k=g\ast(h\ast k)$. 
    \item There exists an $e\in G$ such that $g\ast e=e\ast g=g$ for all $g\in G$.
    \item For each $g\in G$, there exists $g^{-1}\in G$ such that $g\ast g^{-1}=g^{-1}\ast g=e$.
\end{enumerate}
We call $e$ as identity, $g^{-1}$ as inverse of $g$. Furthermore, if $g\ast g'=g'\ast g$ is satisfied for all $g,g'\in G$, we call $G$ as abelian group.\\

Now let $(R,+)$ be a abelian group. If there is one more binary operation $\circ$ on $(R,+)$ satisfying following properties, we call $(R,+,\circ)$ as ring.
\begin{enumerate}
    \item $a\circ(b\circ c)=(a\circ b)\circ c$
    \item $a\circ (b+c)=a\circ b + a\circ c$ for all $a,b,c\in R$.
    \item $(a+b)\circ c=a\circ c + b\circ c$ for all $a,b,c\in R$.
\end{enumerate}
If $a\circ b=b\circ a$ for all $a,b\in R$, we call $(R,+,\circ)$ as commutative ring.\\

Let $R$ be a commutative ring and let's denote $\circ$ as $\cdot$ or abbreviate it. Then we define polynomial ring of $R$ as:
\begin{equation*}
    R[x] := \{a_0+a_1x+a_2x^2+\cdots+a_nx^n:a_i\in R, n\in\mathbb{N}\}
\end{equation*}
Now we define sum between an element $r\in R[x]$ and subset $S\subseteq R[x]$ as follows:
\begin{equation*}
    r + S = \{r+s\in R[x]: s\in S\}
\end{equation*}
And we define ideal $I$ of commutative ring $R[x]$. For every $r\in R[x]$ and every $i\in I$, if $ri\in I$ is satisfied, we call $I$ as ideal of $R[x]$. We denote $I\triangleleft R[x]$ implying $I$ is ideal of $R[x]$.\\

Ideal gives us a new concept, namely quotient ring. Let $R[x]$ be a polynomial ring and $I$ be its ideal. Then
\begin{equation*}
    R[x]/I = \{r+I:r\in R[x]\}
\end{equation*}
we denote elements of $R[x]/I$ as $\bar{r}:=r+I$. Operations between elements of $R[x]/I$, namely $\bar{r}, \bar{s}$ are defined as:
\begin{equation*}
    \bar{r} + \bar{s} = (r+s)+I = \overline{r+s},\quad\bar{r} \cdot \bar{s} = (r\cdot s)+I = \overline{r\cdot s }
\end{equation*}
These operations are mathematically well-defined, and $R[x]/I$ has ring structure as the name quotient ring implies.\\

Now, we denote $(r)$ for smallest ideal containing $r\in R[x]$. Finally, in this paper we only consider the quotient ring $R[x]/(x^2)$ only. We call this quotient ring as dual number ring. The elements of dual number ring has form $a+bx$ where $a,b\in R$ only.\\

\subsection{Properties of Dual Number Ring}
Let $\mathbb{R}$ be ring of real numbers. Then, the dual number ring $R=\mathbb{R}[\varepsilon]/(\varepsilon^2)$ has following property. For every polynomial $f(x)=a_0+a_1x+\cdots+a_nx^n$, following is true:

\begin{align*}
    f(x+\varepsilon) &= a_0 + a_1(x+\varepsilon) + a_2(x+\varepsilon)^2 + \cdots + a_n(x+\varepsilon)^n\\
    &= a_0+a_1x+a_2x^2+\cdots+a_nx^n+\\&\quad\quad\varepsilon(a_1+2a_2x+3a_3x^2+\cdots+na_nx^{n-1})\\
    &=f(x)+f'(x)\varepsilon
\end{align*}
This implies we can perform forward-time differentiation which will be named as forward-propagation.

\subsection{Approximating Non-Polynomial Function}
We want to calculate dual number with non-polynomial function, such as $\exp, \sin, \cos$. This can be performed by using Taylor's series expansion. For example, let's calculate $\sin(x+\varepsilon)$:
\begin{align*}
    \sin(x+\varepsilon) &= (x+\varepsilon) - \frac{(x+\varepsilon)^3}{3!} + \frac{(x+\varepsilon)^5}{5!}-\cdots\\
    &=(x+\varepsilon) - \frac{x^3+3x^2\varepsilon}{3!}+\frac{x^5+5x^4\varepsilon}{5!}-\cdots\\
    &=\bigg(x-\frac{x^3}{3!}+\frac{x^5}{5!}-\cdots\bigg) + \bigg(1-\frac{x^2}{2!}+\frac{x^4}{4!}+\cdots\bigg)\varepsilon\\
    &=\sin x + \varepsilon\cos x
\end{align*}
So we can get differentiation of analytic functions using dual number system.

\section{Learning Rule for Dual Number System}
In this paper, we consider only learning rule for single-layer perceptron with some well-behaved activation function.\\
Let $f$ be a single-layered neural network with $n\times 1$ vector $x=(x_1,\cdots,x_n)^t$ and $1\times n$ weight matrix $W=(w_1,\cdots,w_n)$ and bias $b$, activation function $\sigma$. Then this neural network is formulated as:
\begin{equation}\label{eq1}
    \hat{y}=f(x)=\sigma(Wx+b)
\end{equation}
And let the loss function be $\mathcal{L}:=\mathcal{L}(\hat{y},y)$. Now, let's substitute $x$ to $x+\varepsilon$. Then, (\ref{eq1}) becomes:
\begin{equation}
    f(x+\varepsilon) = \sigma(Wx+b+W\varepsilon)
\end{equation}
For example, let $\sigma$ be sigmoid function:
\begin{equation*}
    \sigma(Wx+b) = \frac{1}{1+\exp(-Wx-b)}
\end{equation*}
And differentiation is:
\begin{equation*}
    \frac{\partial\sigma(Wx+b)}{\partial w_{i}} = x_i\cdot\sigma(Wx+b)(1-\sigma(Wx+b))
\end{equation*}
Therefore,
\begin{equation}
    \frac{\partial\sigma(Wx+b)}{\partial W} = \bigg(\sigma(Wx+b)(1-\sigma(Wx+b))\bigg)\cdot x
\end{equation}
In this section, we will consider only $n=2$ case, namely $x=(x_1,x_2)^t$ and $W=(w_1, w_2)$ case. For higher $n$, the situation is same with $n=2$.
\begin{equation}
    \sigma(Wx+b)=\frac{1}{1+\exp(-x_1w_1-x_2w_2-b)}
\end{equation}
And differentiation is:
\begin{align*}
    \frac{\partial\sigma(Wx+b)}{\partial W}&=\frac{\partial\sigma(x_1w_1+x_2w_2+b)}{\partial(w_1,w_2)}\\
    &=\frac{\partial}{\partial(w_1,w_2)}\bigg(\frac{1}{1+\exp(-x_1w_1-x_2w_2-b)}\bigg)\\
    &=\begin{pmatrix}\sigma(Wx+b)(1-\sigma(Wx+b))x_1&\sigma(Wx+b)(1-\sigma(Wx+b))x_2\end{pmatrix}\\
    &=\sigma(Wx+b)(1-\sigma(Wx+b))\begin{pmatrix}
        x_1&x_2
    \end{pmatrix}
\end{align*}
In the dual number system, calculating $f(x+\varepsilon)=f(x_1+\varepsilon, x_2+\varepsilon)$ is as follows:
\begin{align*}
    f(x_1+\varepsilon, x_2+\varepsilon) &= \sigma(W(x+\varepsilon)+b)\\
    &=\frac{1}{1+\exp(-x_1w_1-x_2w_2-b-(w_1+w_2)\varepsilon)}\\
    &=\sigma(Wx+b)+(w_1+w_2)\cdot\sigma(Wx+b)\cdot(1-\sigma(Wx+b))\varepsilon
\end{align*}
Now let's denote real part and dual part of a dual number $d=a+b\varepsilon$ to be $\mathcal{R}(d)=a$, $\mathcal{E}(d)=b$. Then the above equations have relationship of:
\begin{equation}
    \frac{\partial\sigma(Wx+b)}{\partial W} = \mathcal{E}\bigg(f(x+\varepsilon)\bigg)/(W\cdot\mathbbm{1})\cdot \begin{pmatrix}
        x_1&x_2
    \end{pmatrix}
\end{equation}
Now, set loss function $\mathcal{L}(\hat{y}, y)=(y-\hat{y})^2$ for example. Then, in ordinary chain rule differentiation,
\begin{equation}
    \frac{\partial\mathcal{L}}{\partial(w_1,w_2)}=\frac{\partial\mathcal{L}}{\partial\hat{y}}\frac{\partial\hat{y}}{\partial(w_1,w_2)}=2(\hat{y}-y)\bigg(\sigma(Wx+b)(1-\sigma(Wx+b))\begin{pmatrix}
        x_1&x_2
    \end{pmatrix}\bigg)
\end{equation}
In the dual number system, by denoting $\hat{y}_\varepsilon:=f(x+\varepsilon)$, we can get same result with following rule:
\begin{equation}\label{dualchain}
    2(\mathcal{R}(\hat{y}_{\varepsilon}) - y)\cdot \mathcal{E}(f(x+\varepsilon))/(W\cdot\mathbbm{1})\cdot \begin{pmatrix}
        x_1&x_2
    \end{pmatrix} = 2(\hat{y}-y)\cdot\bigg(\sigma(Wx+b)(1-\sigma(Wx+b))\begin{pmatrix}
        x_1&x_2
    \end{pmatrix}\bigg)
\end{equation}
Getting same result. Therefore, we can perform gradient descent without BP. This flow can be extended in multilayer neural network using chain rule.

\section{Conclusion}
In this paper, we theoretically showed a new framework, named dual number system that performs automatic feed-forward propagation. However, this paper showed only single layered perceptron in a theoretical sense. The multilayer case seems to be solved if we consider chain rule, but the algorithmic details are omitted.
In conclusion, this paper suggests new learning algorithm using dual number system. This new algorithm is fast, light and efficient compared to back-propagation.

\bibliography{ref}


\end{document}